\documentclass[letterpaper, 10 pt, conference]{ieeeconf}  

\IEEEoverridecommandlockouts                              

\usepackage{amssymb}  
\usepackage{flushend}
\usepackage{amsmath}
\usepackage{graphicx}
\usepackage{multirow}
\usepackage{xcolor}
\usepackage{caption}
\usepackage{subcaption}
\usepackage{cite}
\usepackage{float}
\usepackage{booktabs}

\usepackage[capitalize]{cleveref}
\usepackage{array}
\usepackage{tabularx}
\usepackage{siunitx}

\crefformat{figure}{Fig.\hspace{0.33em}#2#1#3}

\usepackage[absolute,showboxes]{textpos}

\TPMargin*{3pt}
\newcommand\copyrighttext{
    \footnotesize
    \noindent
    SUBMITTED TO REVIEW AND POSSIBLE PUBLICATION. COPYRIGHT WILL BE TRANSFERRED WITHOUT NOTICE.\\
    Personal use of this material is permitted.
    Permission must be obtained for all other uses, in any current or future media, including reprinting/republishing this material for advertising or promotional purposes, creating new collective works, for resale or redistribution to servers or lists, or reuse of any copyrighted component of this work in other works.}%
\newcommand\copyrightnotice{%
    \begin{textblock*}{6.6in}(0.95in,0.15in)
        \centering
        \copyrighttext
    \end{textblock*}
}

\title{\LARGE \bf
Dynamic Intent Queries for Motion Transformer-based \\ Trajectory Prediction
}

\author{Tobias Demmler$^{1}$, Lennart Hartung$^{2}$, Andreas Tamke$^{3}$, Thao Dang$^{4}$,\\ Alexander Hegai$^{5}$, Karsten Haug$^{6}$ and Lars Mikelsons$^{7}$
\thanks{$^{1}$Tobias Demmler is with Robert Bosch GmbH, Stuttgart, Germany
        {\tt\small tobias.demmler@de.bosch.com}}%
\thanks{$^{2}$Lennart Hartung is with Robert Bosch GmbH, Stuttgart, Germany
        {\tt\small lennart.hartung@de.bosch.com}}%
\thanks{$^{3}$Andreas Tamke is with Robert Bosch GmbH, Stuttgart, Germany
        {\tt\small andreas.tamke@de.bosch.com}}%
\thanks{$^{4}$Thao Dang is with the Institute for Intelligent Systems, Department of Computer Science and Engineering, Esslingen University, Germany
        {\tt\small thao.dang@hs-esslingen.de}}%
\thanks{$^{5}$Alexander Hegai is with Robert Bosch GmbH, Stuttgart, Germany
        {\tt\small alexander.hegai@de.bosch.com}}%
\thanks{$^{6}$Karsten Haug is with Robert Bosch GmbH, Stuttgart, Germany
        {\tt\small karsten.haug@de.bosch.com}}%
\thanks{$^{7}$Lars Mikelsons leads the Chair of Mechatronics of the University of Augsburg, Germany
        {\tt\small lars.mikelsons@uni-a.de}}%
}

\begin{document}
\copyrightnotice

\maketitle
\thispagestyle{empty}
\pagestyle{empty}

\begin{abstract}

In autonomous driving, accurately predicting the movements of other traffic participants is crucial, as it significantly influences a vehicle's planning processes. Modern trajectory prediction models strive to interpret complex patterns and dependencies from agent and map data. The Motion Transformer (MTR) architecture and subsequent work define the most accurate methods in common benchmarks such as the Waymo Open Motion Benchmark. The MTR model employs pre-generated static intention points as initial goal points for trajectory prediction. However, the static nature of these points frequently leads to misalignment with map data in specific traffic scenarios, resulting in unfeasible or unrealistic goal points. Our research addresses this limitation by integrating scene-specific dynamic intention points into the MTR model. This adaptation of the MTR model was trained and evaluated on the Waymo Open Motion Dataset. Our findings demonstrate that incorporating dynamic intention points has a significant positive impact on trajectory prediction accuracy, especially for predictions over long time horizons. Furthermore, we analyze the impact on ground truth trajectories which are not compliant with the map data or are illegal maneuvers.

\end{abstract} 
\section{Introduction}

\begin{figure}
    \vspace{1.5mm}
    \centering
    \fbox{
        \includegraphics[trim=50 250 420 300, clip, width=0.65\columnwidth]{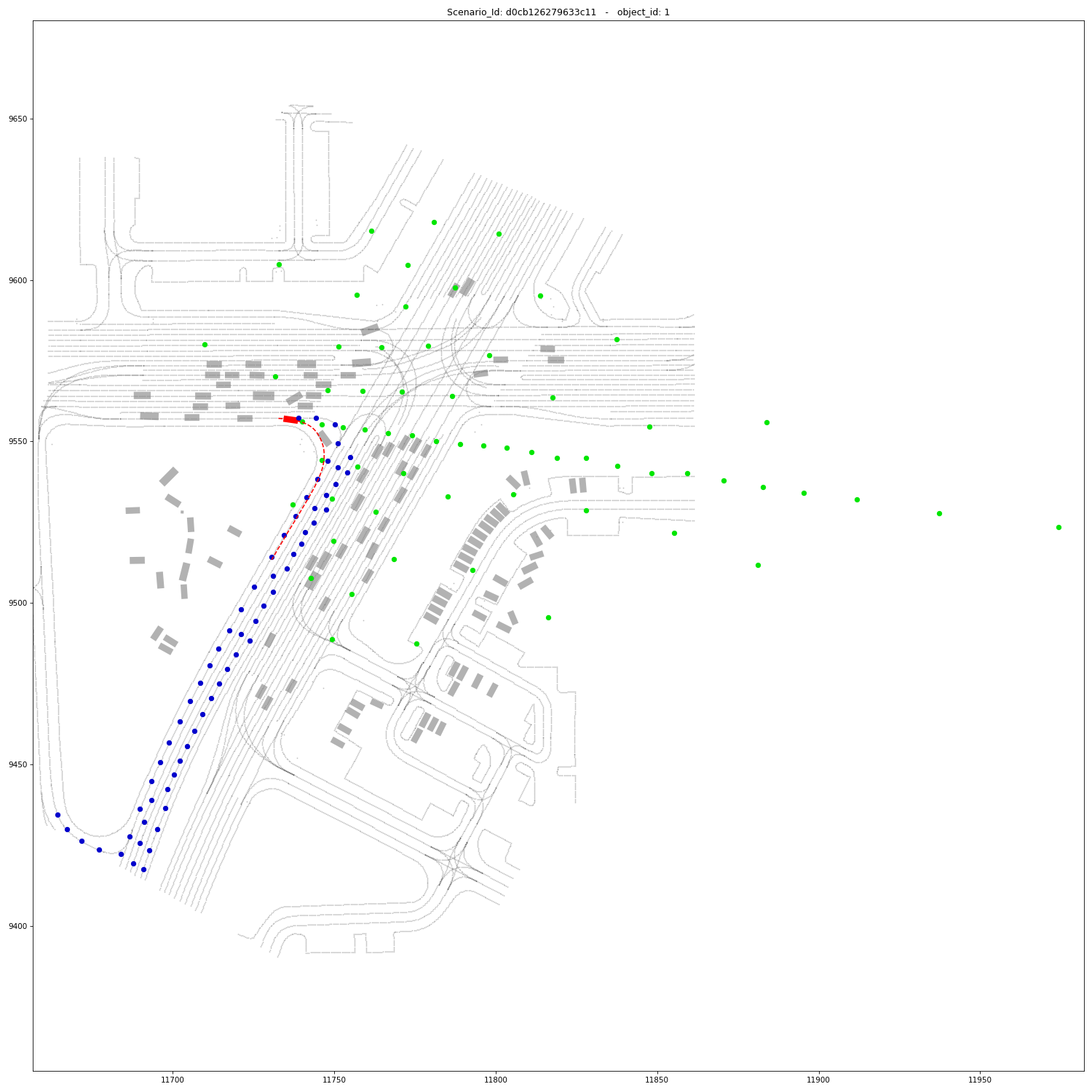}
    }
    \caption[Comparison of Static and Dynamic Intention Points]
    {Comparison of static (green) and dynamic (blue) intention points. The agent (red box) shows its past 1s and future 8s trajectory (red line).
    }
    \label{fig:staticintentroad}
    \vspace{-2mm}
\end{figure}

Trajectory prediction is crucial for modern autonomous driving systems. It forms a deeper understanding of how other traffic participants will move in the future, which is the basis for subsequent motion planning of the autonomous vehicle. Trajectory prediction requires historical agent data of other traffic participants and for many state-of-the-art approaches also a high-definition map (HD map). Different benchmarks are commonly used for the comparison of different approaches, e.g. Argoverse 1 and 2 \cite{Argoverse, Argoverse2} and the Waymo Open Motion Dataset (WOMD)~\cite{ettinger2021large}. An outstanding approach, which is a top performer on these benchmarks, is the Motion Transformer architecture (MTR)~\cite{shi2023mtr}. One key aspect of this architecture is the use of static motion queries. These are points that are statistically viable trajectory endpoints that are used to query the transformer decoder in the network. However, the static nature of these points often leads to misalignment with specific traffic scenarios, resulting in unfeasible or unrealistic initial trajectory targets. In the original MTR paper, the authors compare their static intention points against uniformly distributed intention points. The uniformly distributed intention points are sampled in an 8\texttimes8~grid over the feasible range of trajectories for each object class. The static intention points resulted in a significant improvement across all metrics~\cite{shi2023mtr}. Inspired by these improvements, we aim to further enhance MTR's capabilities by refining these motion queries by generating scene-specific dynamic intention points. See \cref{fig:staticintentroad} for a visual comparison between static and dynamic intention points.

Thus, our contributions in this paper are twofold:
\begin{itemize}
    \item Firstly, we introduce a model for dynamically sampled intention points and an additional model using a mixture of dynamic and static intention points. These models are evaluated and compared against the original static intention point model on the WOMD benchmark. We observe improved performance for the vehicle class. Surprisingly, also the pedestrian and cyclist objects achieve better results despite them not directly using dynamic intention points. 
    
    \item Secondly, we provide an in-depth analysis of how intention point determination is related to legal driving and how that impacts trajectory prediction. This demonstrates that our dynamic approach excels especially in legal maneuvers.
\end{itemize}

\section{Related Work}

Most trajectory prediction frameworks can be classified by two key aspects: First, the encoder architecture which is used to encode the historical agent states and corresponding HD map. Second, the modeling of the multimodal future trajectories.

\subsection{Input Encoding}\label{sec:input_encoding}
For precise trajectory predictions, it is imperative for models to adeptly encode information and understand the interplay between the agent's states and the encompassing road map. Several approaches have been explored for encoding road maps. Early research predominantly employed a rasterization-based methodology using Convolutional Neural Networks (CNNs), where the world's state is depicted as a stack of top-down images~\cite{park2020diverse, marchetti2021mantra, casas2019spatiallyaware}. However, this approach has limitations, such as difficulties in modeling extensive-range interactions, a restricted field of view, and challenges in capturing continuous physical states. Moreover, CNNs require significant computational resources, and the rasterization structure leads to unnecessary calculations for areas of minimal interest. High-resolution representations of relevant zones necessitate increased resolution for the entire map.

A departure from this methodology is observed in VectorNet~\cite{gao2020vectornet}, which represents map elements and agent trajectories as separate polylines, which are then connected in a graph and processed by a Graph Neural Network (GNN). This technique, adopted in recent works~\cite{varadarajan2021multipath, ngiam2022scene, shi2022motion, Zhou2023QueryCentricTP}, capitalizes on the inherent sparsity of road networks and their concise representation as polylines. LaneGCN~\cite{liang2020learning} distinguishes itself by conceptualizing road lanes as nodes within a GNN, leveraging the structural connectivity of road networks. This technique was adopted by more recent works \cite{gilles2021thomas, Demmler2024TowardsCA}.

A critical consideration in motion history modeling is the chosen frame of reference for encoding. Some methodologies opt for a global, scene-level coordinate frame, as seen in the rasterized top-down image approach. This method offers efficiency through a singular, unified representation of the world state but may compromise pose-invariance. Conversely, models operating within the agent-coordinate frame inherently provide pose-invariance but increase computational demands linearly with the number of agents, or even quadratically considering pairwise interactions. A promising approach is proposed by GoRela~\cite{Cui2022GoRelaGR}. Here the positional encoding is not defined by a specific coordinate but by the distance and orientation to other scene elements. This approach is extended to simultaneous prediction and planning in \cite{hagedorn2024pioneering}.

Given the advantages and disadvantages of these approaches, many research efforts adopt a hybrid approach. Works such as those by Shi et al. (2023), Ngiam et al. (2022), and Liang et al. (2020)~\cite{shi2022motion, ngiam2022scene, liang2020learning} illustrate this. They present agent-centric models residing within a global frame, enabling efficient analytical reasoning. This strategy also captures distant interactions and preserves detailed state representations while avoiding the limitations of rasterization.

\subsection{Multimodal Future Behavior Modeling}\label{sec:behavior_modeling} Given the encoded scene context features, contemporary research employs a variety of strategies for simulating an agent's potential future actions in a multimodal context. Early works~\cite{gupta2018social, Rhinehart_2018_ECCV} advocate the generation of multiple trajectory forecasts to capture the range of possible outcomes. Meanwhile, studies such as Chai et al. (2019) and Mercat et al.~\cite{chai2019multipath, mercat2019multihead} have proposed using Gaussian Mixture Models (GMMs) to encapsulate multimodal forecasts in a more compact probabilistic framework. An alternative approach is found in goal-oriented methods~\cite{gilles2021gohome, gu2021densetnt, zhao2020tnt}, which initially determine the high-level intentions of the agent before delineating specific path predictions. Their approach involves the use of a dense selection of goal candidates to cover all potential agent destinations. While the use of these goal candidates reduces the optimization workload by decreasing trajectory ambiguity, their density introduces a trade-off: Reducing the number of candidates compromises performance while increasing the number of candidates escalates computational and memory demands.

On the other hand, direct-regression methods, as referenced in~\cite{ngiam2022scene,varadarajan2021multipath}, predict trajectories directly from the encoded features of an agent, ensuring an adaptive representation of the agent's future behavior. While they offer the advantage of flexibility, accommodating a wide spectrum of agent behaviors, these methods often suffer from slow convergence. This is mainly due to the need to regress diverse motion modes from a singular agent feature, without the benefit of employing spatial priors. Additionally, these methods tend to predict dominant modes prevalent in the training data, as these modes significantly influence the optimization of the agent feature.

The Motion Transformer (MTR)~\cite{shi2022motion} combines these two types of methods. It achieves this fusion by adopting a modest set of static intention points. This set is generated by reducing all observed trajectory endpoints in the dataset to 64 points via K-means clustering. These scene independent static intention points are used to query the transformer model for a diverse set of trajectory predictions. This approach eliminates the necessity for an extensive set of goal candidates and mitigates optimization difficulties. 

Later versions of MTR, namely MTR++~\cite{shi2023mtr} and MTR\,v3~\cite{shi2024mtrv3}, improve upon the original MTR architecture. We decided to use the original MTR architecture for several reasons. Firstly, only the implementation for the original MTR architecture is publicly available. By adding the modifications mentioned in the corresponding papers for the later versions we would risk adding unintentional deviations from the original unpublished implementations. Secondly, the improvements of MTR++ do not modify the intention point aspect of the architecture and therefore our findings with the original architecture directly translate to this version. Lastly, in MTR\,v3 the authors incorporate the findings of "Evolving and Distinct Anchors" (EDA)~\cite{lin2024eda}. Here the network learns to iteratively shift the static intention points based on the current scene. However, they still use the static intention points as a starting point for each prediction. Our dynamic intention points can improve the initial intention points used in the EDA approach and therefore also enhance the MTR\,v3 model. Therefore, using the original MTR implementation allows us to achieve the most comparable results which also translate to later versions.

A notable adaptation of the MTR framework is the ControlMTR model~\cite{Sun2024ControlMTRCM}, which employs traffic-rule-compliant intention points. These points are similar to our dynamic intention points. While ControlMTR prioritizes the development of its novel encoder-decoder architecture to enhance trajectory prediction, our work provides a systematic investigation into the explicit influence of dynamic intention points, with an emphasis on modeling illegal maneuvers and their implications for real-world driving scenarios. Crucially, our analysis of illegal maneuvers is also applicable to ControlMTR’s framework.
   
\section{Problem Definition}
\begin{figure}
    \vspace{1.5mm}
    \centering
    \includegraphics[width=\columnwidth]{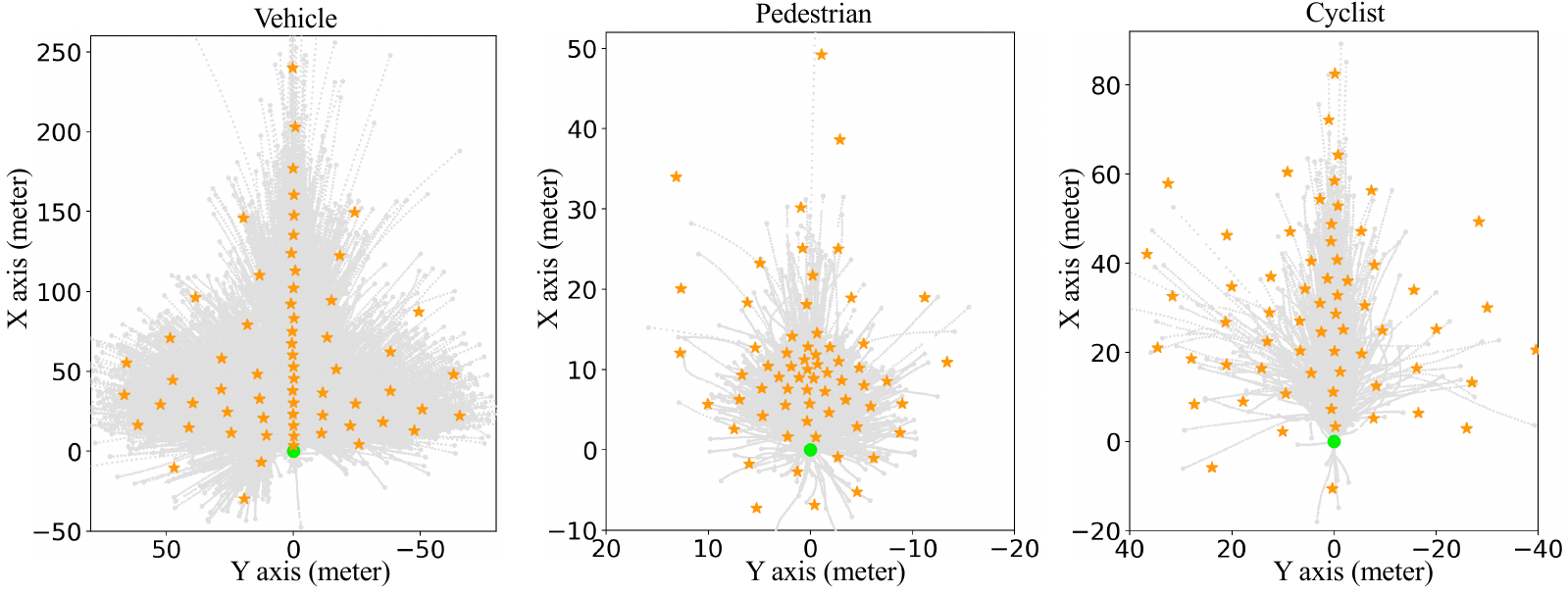}
    \caption[Vehicle Static Intention Points Distribution]
    {Static intention point distributions for vehicles, pedestrians and cyclists. Agent's current position (green dot), intention points (orange stars), and for clearer visualization only 10\% of the historical ground-truth trajectories (gray dotted lines). Taken from~\cite{shi2022motion}.
    }
    \label{fig:mtr_staticintents}
\end{figure}

The MTR model distinguishes itself from other motion prediction models primarily through its use of motion queries. These queries employ static intention points to forecast an entity's prospective destination. Each intention point serves as an anchor, facilitating the progressive refinement of trajectory predictions. These points are derived using a K\nobreakdash-Means clustering algorithm, which analyzes the ground-truth trajectory endpoints from the training dataset. This algorithm groups similar endpoints together, forming clusters that represent common destinations or paths. This process is applied separately for different entities such as vehicles, pedestrians, and cyclists. The outcome for these object classes is illustrated in \cref{fig:mtr_staticintents}, showing 64 static intention points for each class.

A critical aspect to consider is that these static intention points are initially generated without considering the constraints of the road network. Consequently, a substantial number of these points do not correspond to feasible or realistic trajectory goal points, as depicted in \cref{fig:staticintentroad}.

To enhance the accuracy of trajectory predictions, we introduce a dynamic approach that leverages road map data to refine the generation of intention points. This refinement aims to reduce the prevalence of infeasible trajectory endpoints, as seen in \cref{fig:staticintentroad}. This methodology particularly addresses vehicle trajectories, given their more predictable, lane-bound movement patterns. This contrasts with the more erratic and lane-independent movements of pedestrians and cyclists. These latter groups also appear less frequently in the dataset and present a higher complexity in their movement behaviors. Consequently, our experimental focus is on refining vehicle trajectory predictions by integrating road map information into the intention point generation process. This integration improves the precision and realism of trajectory forecasts.

\section{Generation of Dynamic Intention Points}\label{sec:dynamic_intents}

In order to generate our dynamic intention points we have to first associate each agent with a specific lane. Based on this lane association we can then extrapolate all legal movements into a road graph. Lastly, we have to condense this road graph to our dynamic intention points.

\subsection{Lane Association}\label{sec:vehicle_local}
Accurately assigning a specific lane within the HD map to a vehicle is fundamental to making further movement predictions. A straightforward method to determine a vehicle's specific lane and position within that lane is to identify the nearest lane node relative to the vehicle's position. This involves calculating the Euclidean distance from the vehicle's center to all designated vehicle lane nodes. The lane node exhibiting the shortest distance is then considered to represent the lane in which the vehicle is currently traveling. However, this approach has limitations. There can be multiple overlapping lane segments in the HD map. Particularly in complex situations such as intersections. Furthermore, drivers do not always adhere strictly to the lanes. Such discrepancies can lead to certain edge cases that must be addressed. If the handling of these edge cases does not result in a proper lane association we fall back to the static intention points for that specific agent. 

\begin{figure}
    \vspace{1.5mm}
    \begin{subfigure}[b]{0.295\columnwidth}
        \fbox{\includegraphics[trim=550 350 50 250, clip, width=\columnwidth]{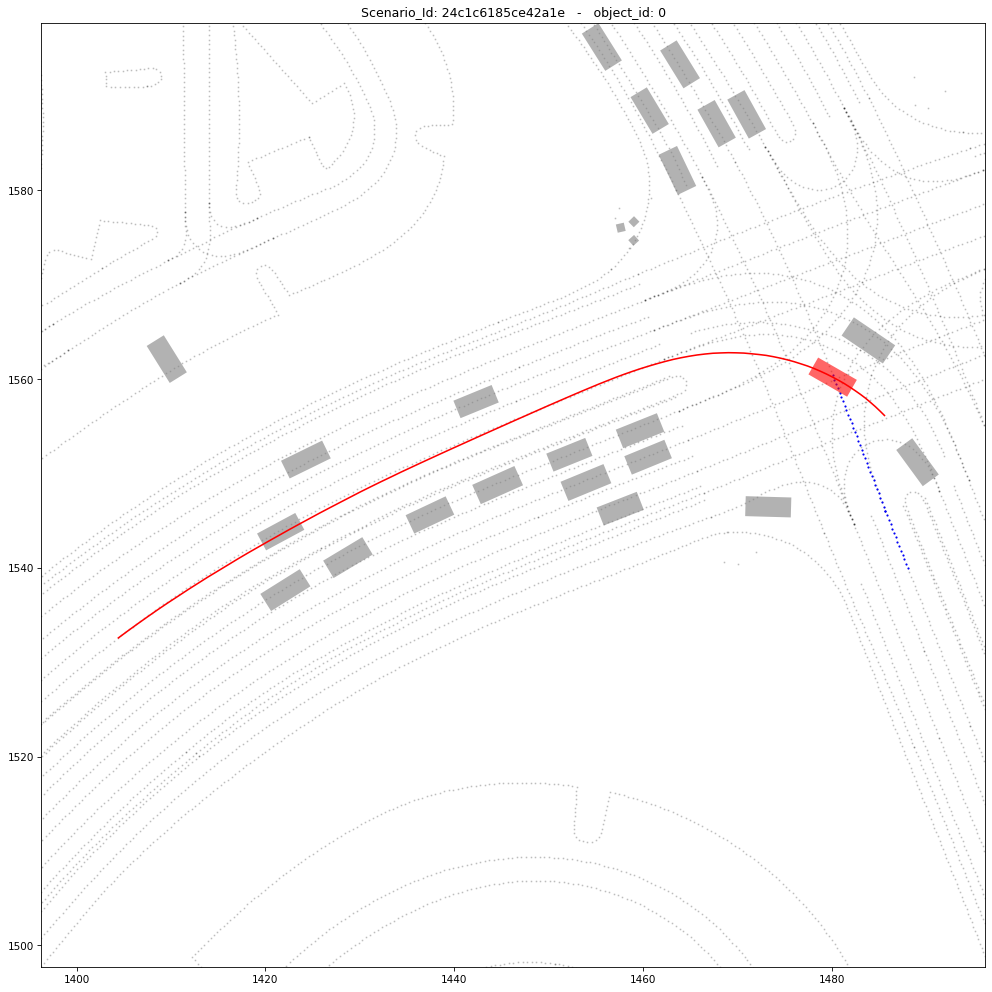}}
        \caption{}
        \label{edge_case:a}
    \end{subfigure}
    \hspace{3pt}
    \begin{subfigure}[b]{0.295\columnwidth}
        \fbox{\includegraphics[trim=300 300 300 300, clip, width=\columnwidth]{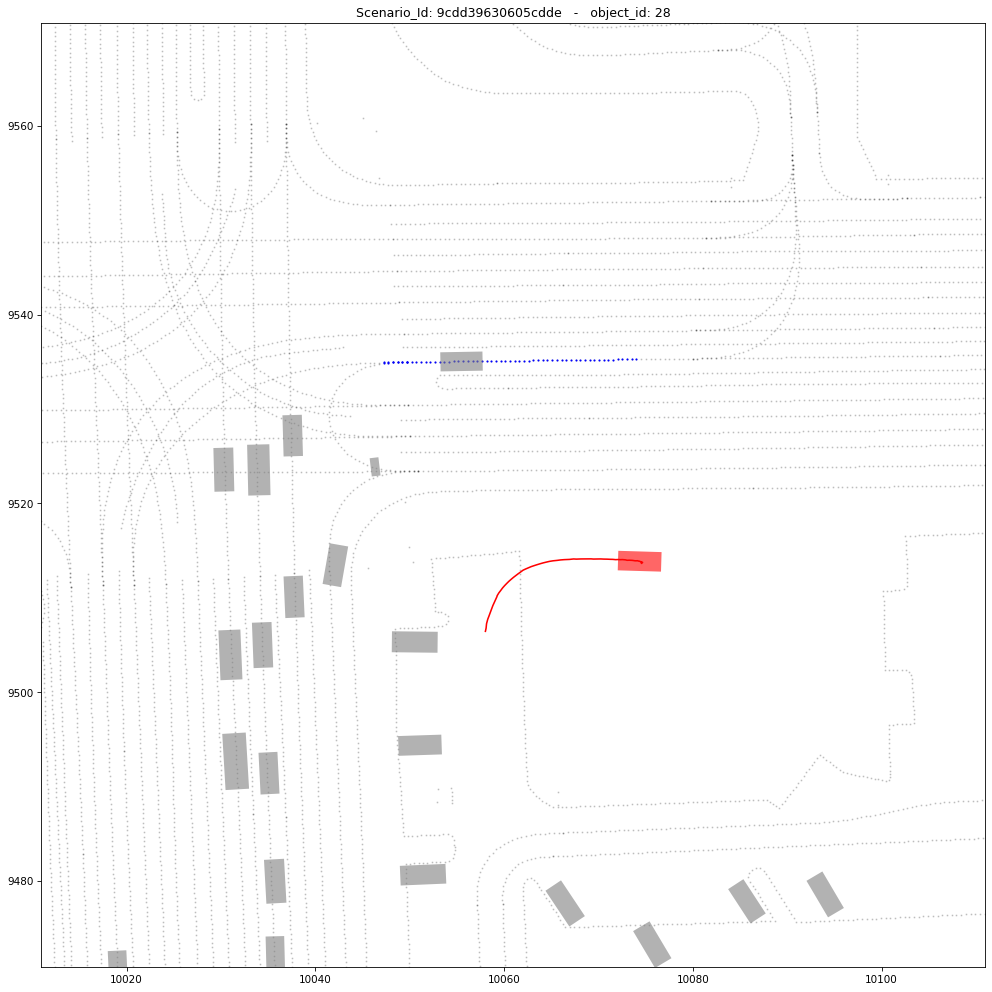}}
        \caption{}
        \label{edge_case:b}
    \end{subfigure}
    \hspace{3pt}
    \begin{subfigure}[b]{0.295\columnwidth}
        \fbox{\includegraphics[trim=550 350 100 300, clip, width=\columnwidth]{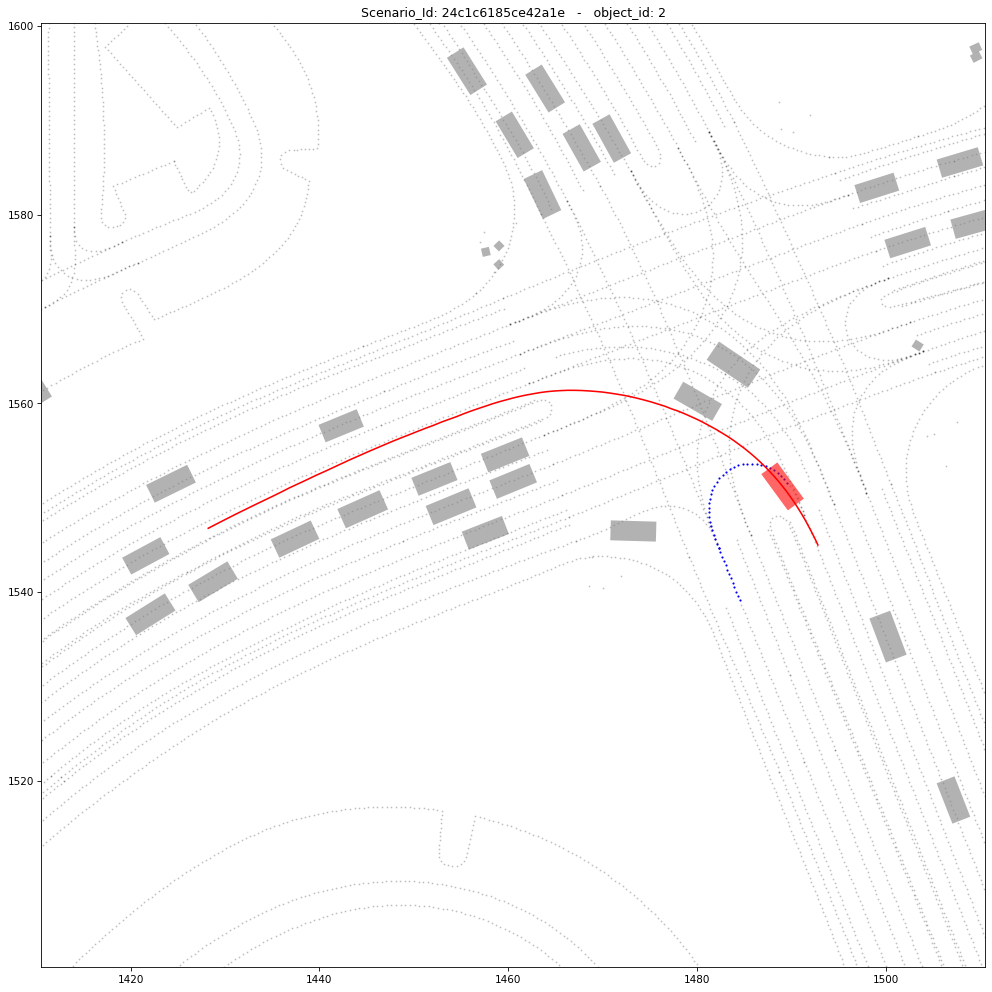}}
        \caption{}
        \label{edge_case:c}
    \end{subfigure}
    \caption[Edge Cases in Vehicle Localization]
    {Edge cases in scenarios without validity checks: Heading Alignment (a), Proximity Limit (b), and 'Backwards Look' (c). The target agent is highlighted in red, with its actual past and future trajectory shown by a red line and the incorrectly assigned lane indicated by blue marked lane nodes.
    }
    \label{fig:edge_case}
    \vspace{-4mm}
\end{figure}

\textbf{Heading alignment between agent and lane:}
One notable edge case occurs during intersection navigation or instances of corner cutting during a turn maneuver. A vehicle traversing orthogonal lanes or cutting corners during a turn may momentarily align closer to a node of an intersecting lane than to its actual lane, as seen in \cref{edge_case:a}. To mitigate this, we introduce a validation step assessing the alignment between the vehicle's and the lane's headings within a pre-established threshold. The lane's heading is calculated by the angle between two successive points on the lane segment, while the vehicle's heading is derived from its state history. A minimum threshold angle of 45 degrees is set for alignment verification, effectively addressing misalignments at intersections and allowing for a substantial degree of tolerance during typical driving maneuvers such as turns, lane changes, or instances of corner cutting.

\textbf{Proximity limit for off-road scenarios:}
In situations where a vehicle operates in a parking lot adjacent to a street, the absence of map data for these areas poses a challenge. Without a proximity limit from the vehicle to the lane node, situations like those shown in \cref{edge_case:b} will arise. To address this, we set a maximum allowable deviation of 5m from a recognized lane for the vehicle to remain eligible for lane-based prediction. This distance accommodates substantial deviations, including extensive corner-cutting or off-road driving, while preserving realistic on-road navigation boundaries. If a vehicle is found in a scenario where no lane satisfies this criterion, the scene is considered unsuitable for generating dynamic intents. In such cases, we default to using static intents, ensuring that predictions remain grounded in realistic driving scenarios, even when comprehensive map data is lacking.

\textbf{Overlapping lane segments:}
Scenarios involving multiple overlapping lane segments present a unique edge case. For example, consider a street where one of the left lanes is designated for U-turns alongside regular left turns. In such a setup, as displayed in \cref{edge_case:c}, a vehicle executing a left turn but cutting the corner might be inaccurately assigned to the U-turn lane segment. To address this issue, we have incorporated a 'backwards look' logic. This logic extends 10m back along the lane to which the vehicle is currently assigned. In situations where a single lane segment diverges into multiple segments, such as a split for U-turns and left turns, the nearest lane node of each emerging lane segment is identified. If these nodes fall within the proximity limit set out in the previous paragraph, they are considered as potential correct assignment points.

\subsection{Road graph generation}\label{sec:roadgraph}
After assigning a vehicle to a specific lane we need to generate a road graph containing all subsequently reachable positions. The road graph generation begins from the previously identified starting positions, using the Dijkstra algorithm~\cite{dijkstra1959note}.

In creating the weighted graph from polyline inputs, we transform lane nodes extracted from polylines into corresponding graph nodes. This transformation necessitates establishing weighted connections among these lane nodes. Primarily, connections are formed between consecutive points within a lane segment. Additionally, we link the end node of each lane segment to the start node of an adjacent exit lane and create reciprocal links for entry lanes. Given the frequent occurrence of multiple adjacent lanes, it is essential to also consider lane changes if permitted by lane markings. This necessitates identifying the nearest node on neighboring lanes relative to a given node and incorporating this node as a reachable point in the graph.

Our implementation's optional termination criterion is the maximum permissible travel time to a node, set at 8 seconds, corresponding with our maximum prediction horizon. The weight assigned to each graph edge signifies the travel time between nodes, computed using the Euclidean distance and the speed limit at the originating node. Acknowledging that actual driving speeds frequently surpass posted limits, we adjust our speed computations by adding an extra 15 mph. This increment is based on the highest speed limit violation observed in the dataset. This gives us a set of points that is guaranteed to include all map conform maneuvers. We could add additional constraints such as current speed and maximum acceleration or deceleration, the appropriate speed for a turn might be lower than the speed limit, or maybe there is a traffic jam. We elected to dismiss these additional constraints as they introduce additional complexity and the neural network is better suited to handle these increasingly complex interactions. 

\begin{figure}
    \vspace{1.5mm}
    \centering
    \fbox{\includegraphics[trim=175 25 50 25, clip, width=0.6\columnwidth]{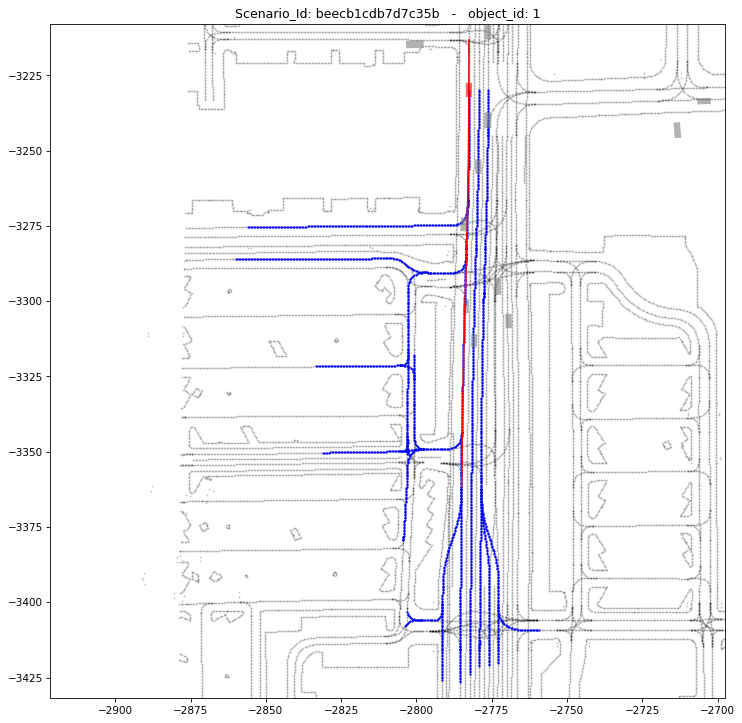}}
    \caption[Illustration of a Generated Road Graph]
    {A visual depiction of a generated road graph, marked in blue. The agent and its corresponding ground-truth trajectory are illustrated in red.}
    \label{fig:roadgraph}
\end{figure}

The road graph is thus constituted by all points that are accessible from the initial position within an 8 second time frame, as illustrated in \cref{fig:roadgraph}. This approach guarantees that our road graph comprehensively encompasses the feasible range of vehicular movement, taking into account both the constraints imposed by traffic regulations and the specifications provided by the HD map.

\subsection{Point-sampling}\label{sec:pointSampling}
\begin{figure}
    \vspace{1.5mm}
    \centering
    \fbox{\includegraphics[trim=50 100 10 50, clip, width=0.71\columnwidth]{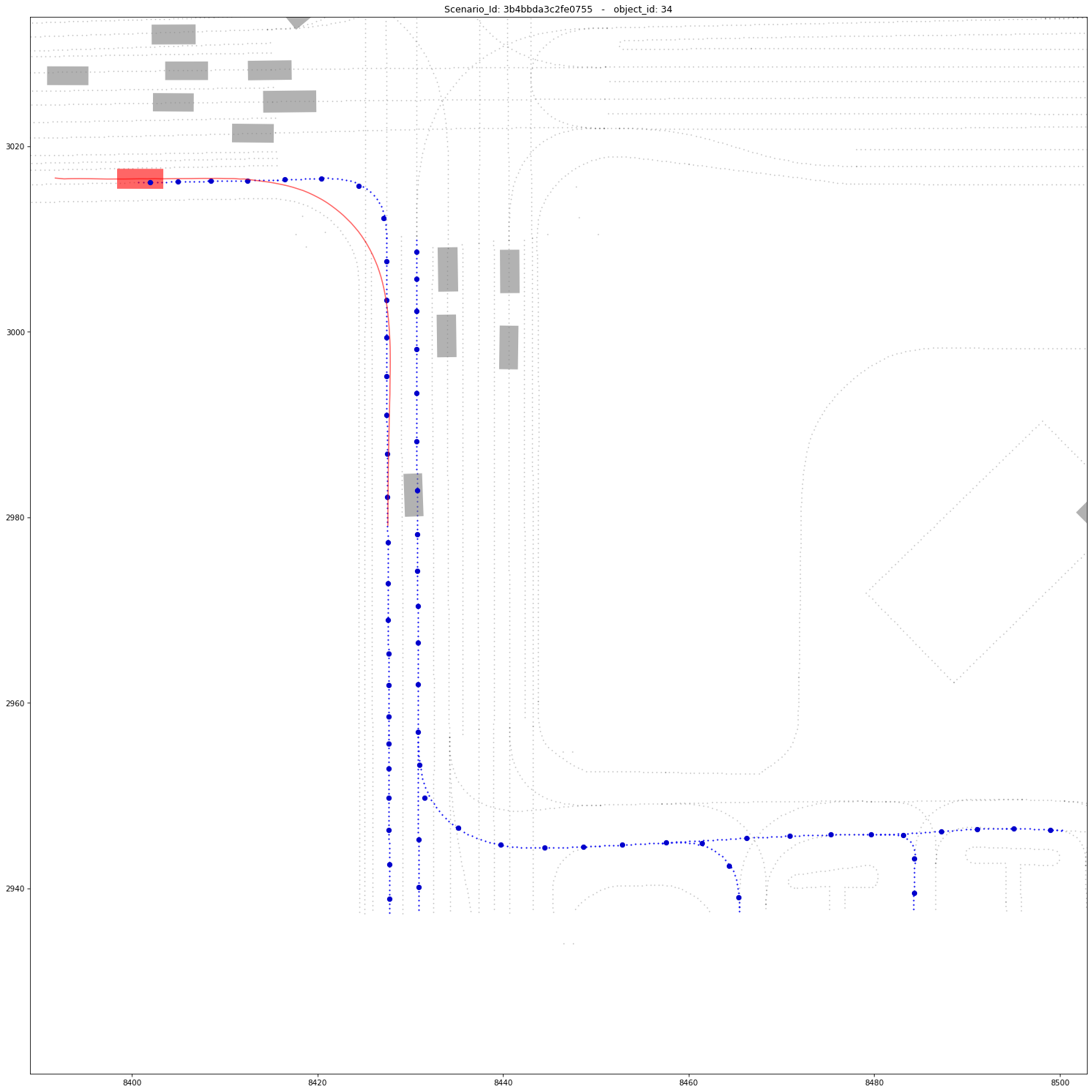}}
    \caption[Point-sampling on a Road Graph]
    {Intention points (blue dots) on the road graph (blue line) generated via K-Means Clustering, with the agent and its ground-truth trajectory shown in red.
    }
    \label{fig:sampling}
    \vspace{-3mm}
\end{figure}

Given the extensive number of points generated in the road graph, direct input into the transformer model presents a computational challenge. To address this issue, we follow MTR and reduce this point set to 64 points using K-Means clustering. A visual example of this sampling method is illustrated in \cref{fig:sampling}.

\subsection{Hybrid Approach for Illegal Maneuvers}
\begin{figure}
    \vspace{1.5mm}
    \centering
    \fbox{\includegraphics[trim=50 400 420 400, clip, width=0.97\columnwidth]{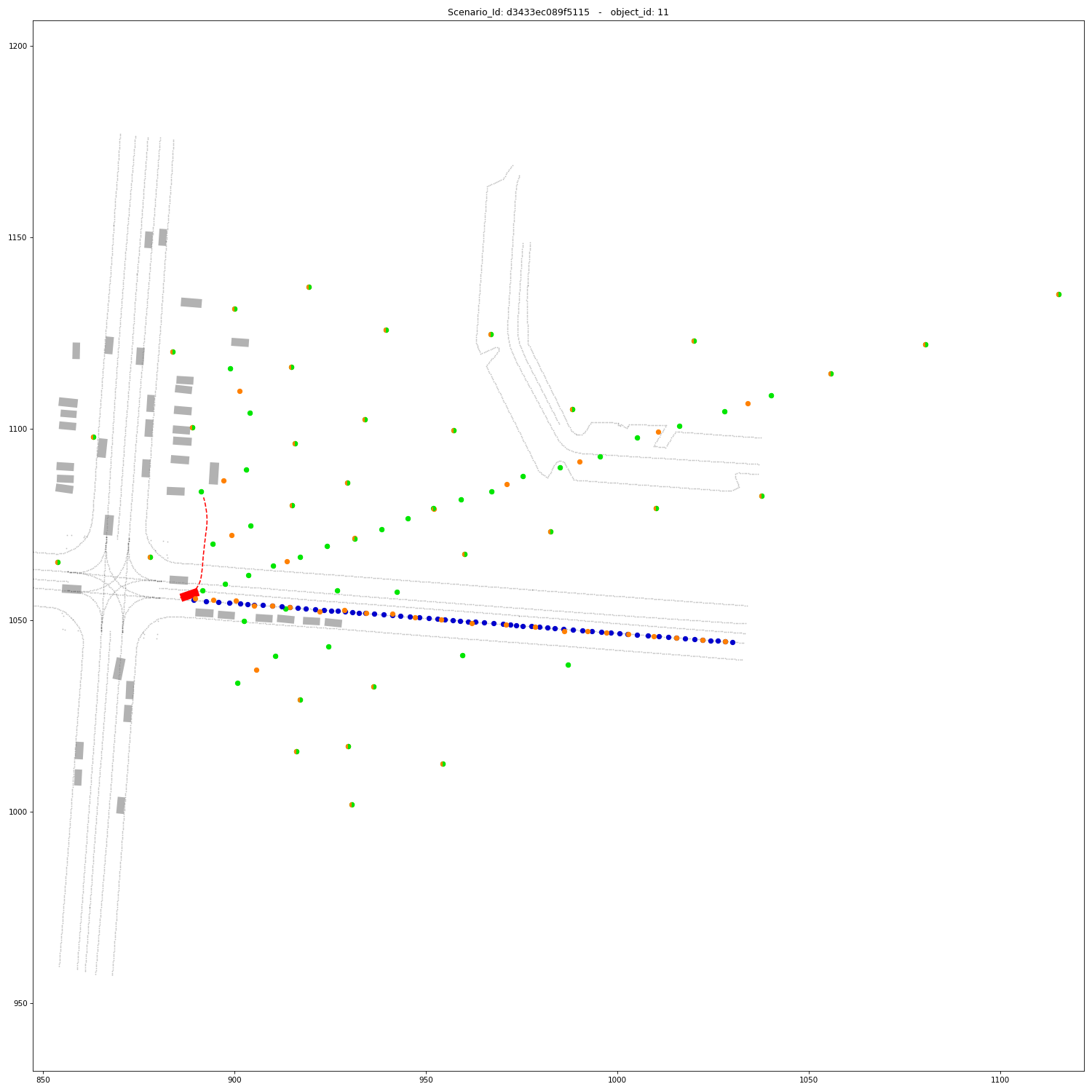}}
    \caption[Traffic Scenario Comparison for Intention Points]
    {Comparison between dynamic intention points (blue) and static intention points (green) and mixed intention points (orange). The agent, and its ground truth path is red.
    }
    \label{fig:illegally_moves}
    \vspace{2mm}
\end{figure}

A key limitation of the dynamic approach is the potential omission of illegal or non map compliant maneuvers not represented in the road graph, as illustrated in \cref{fig:illegally_moves}. To mitigate this, we propose a hybrid approach combining static and dynamic intention points. For this approach, we pool the dynamic intention points with the static intention points and reduce this larger set again to 64 points using K-Means Clustering. This fusion ensures that the intention points emphasize the goal points of legal movements but still provide the flexibility of the static intention points regarding non map conform behavior. To emphasize the dynamic intention points in the mixed model we weigh them at a 3:1 ratio in the clustering process. We evaluated multiple ratios and found this ratio to provide the best performance as can be seen in \cref{tab:combination_distr}.

\begin{table}[b]
    \centering
    \begin{tabular}{cc}
        \toprule
        Ratio (dynamic:static) & minFDE 8s $\downarrow$ \\ \midrule
        1:1                    & 3.024                  \\
        3:1                    & \textbf{3.019}                  \\
        5:1                    & 3.022                  \\
        \bottomrule
    \end{tabular}
    \caption[Influence of Different Intent Ratios on minFDE 8s]
    {Performance analysis of mixed model using various static to dynamic intention point ratios in K-Means Clustering for intention point generation.}
    \label{tab:combination_distr}
\end{table}

\section{Training}\label{sec:trainingsprocess}
Due to memory constraints, we used an optimized model with a reduced number of decoder layers (3 instead of 6~layers) and a reduced batchsize (40 instead of 80 samples). This modification had a negligible impact on the model's performance on the WOMD benchmark, as detailed in \cref{tab:decoder_layer}.

\begin{table}[h]
    \centering
        \begin{tabularx}{\linewidth}{@{\hspace{2pt}}c@{\hspace{8pt}}ccccc}
            \toprule
            Dec. Layers & Batchsize & minADE $\downarrow$ & minFDE $\downarrow$ & MR $\downarrow$ &  mAP $\uparrow$ \\ \midrule
            3         & 40             & 0.6021              & 1.2258              & 0.1335                                   & 0.4192         \\
            6 & 80 & 0.6046 & 	1.2251 	& 0.1366  & 0.4164 \\
            \bottomrule
        \end{tabularx}
    \caption[Impact of the Number of Decoder Layers on Model Performance]
    {Performance analysis of the number of decoder layers and batchsize on the static model on the WOMD dataset. The last row shows the original results for MTR as published in \cite{shi2022motion}.}
    \label{tab:decoder_layer}
    \vspace{-1mm}
\end{table}
   
\section{Experiments}\label{chap:experiments}

We evaluate the three previously mentioned variants of the network: The original static model, the dynamic model, and the mixed model. Each model was trained up to epoch 40, with the model achieving the highest mAP metric at
this juncture selected for further evaluation.
The evaluation is done utilizing the official benchmark metrics detailed in the WOMD paper and competition website \cite{ettinger2021large, WOMDbench2024}. 

\subsection{Results of trained models}

\begin{table}
    \vspace{1.5mm}
    \centering
    \begin{tabularx}{7.25cm}{lcccc}
        \toprule
                Model & 
                {minADE $\downarrow$} & 
                  {minFDE $\downarrow$} & 
                  {MR $\downarrow$} &  
                  {mAP $\uparrow$}  \\ 
        \midrule
        Static  & 0.6021           & 1.2258           & 0.1335            & 0.4192           \\
        Dynamic & 0.5923           & \bfseries 1.1947 & \bfseries 0.1332  & 0.4154           \\
        Mixed   & \bfseries 0.5895 & 1.1962           & 0.1351            & \bfseries 0.4197 \\
        \bottomrule
    \end{tabularx}
    \caption{Performance comparison of motion prediction on the validation set of Waymo Open Motion Dataset.}
    \label{tab:metric_overall}
\end{table}

\def\spacing{7.8}
\begin{table*}
    \vspace{1.5mm}
    \centering
    \begin{tabularx}{0.98\linewidth}{ll|ccc|ccc|ccc|ccc}
        \toprule
        & & \multicolumn{3}{c|}{minADE $\downarrow$} 
        & \multicolumn{3}{c|}{minFDE $\downarrow$} 
        & \multicolumn{3}{c|}{MR $\downarrow$} 
        & \multicolumn{3}{c}{mAP $\uparrow$} \\
        Obj. Class & Model & {3s} & {5s} & {8s} & {3s} & {5s} & {8s} & {3s} & {5s} & {8s} & {3s} & {5s} & {8s}  \\ 
        
        \midrule
        Vehicle     
        & Static  & \bfseries 0.3353 & \bfseries 0.6733 & 1.2541            &  
                  \bfseries 0.6102 & 1.3237           & 2.6326            &  
                  \bfseries 0.1120 & \bfseries 0.1485 & 0.1979            &  
                  \bfseries 0.5412 & \bfseries 0.4494 & 0.3724               
        \\
        & Dynamic & 0.3645           & 0.7069           & 1.2687            &  
                  0.6580           & 1.3749           & 2.5664            &  
                  0.1320           & 0.1537           & \bfseries 0.1655  &  
                  0.4767           & 0.4205           & 0.3721               
        \\ 
        & Mixed   & 0.3420           & 0.6759           & \bfseries 1.2385  &  
                  0.6126           & \bfseries 1.3175 & \bfseries 2.5466  &  
                  0.1137           & 0.1486           & 0.1932            &  
                  0.5189           & 0.4465           & \bfseries 0.3771     
        \\
        \midrule           
        Pedestrian  
        & Static  & \bfseries 0.1627 & 0.3151           & 0.5568            &  
                  0.3074           & 0.6383           & 1.2232            &  
                  0.0592           & 0.0696           & 0.0896            &  
                  0.4728           & 0.3990           & 0.3708               
        \\
        & Dynamic & 0.1647           & \bfseries 0.3146 & \bfseries 0.5511  &  
                  \bfseries 0.3067 & \bfseries 0.6313 & \bfseries 1.2001  &  
                  \bfseries 0.0553 & 0.0673           & \bfseries 0.0847  &  
                  0.4941           & 0.4339           & 0.4015               
        \\
        & Mixed   & 0.1676           & 0.3194           & 0.5597            &  
                  0.3127           & 0.6417           & 1.2110            &  
                  0.0569           & \bfseries 0.0641 & 0.0863            &  
                  \bfseries 0.5045 & \bfseries 0.4404 & \bfseries 0.4207     
        \\
        \midrule
        Cyclist  
        & Static  & 0.3451           & 0.6469           & 1.1294            &  
                  0.6343           & 1.2815           & 2.3811            &  
                  0.1772           & \bfseries 0.1633 & \bfseries 0.1842  &  
                  0.4143           & \bfseries 0.4007 & \bfseries 0.3524     
        \\
        & Dynamic & \bfseries 0.3199 & \bfseries 0.5910 & \bfseries 1.0493  &  
                  \bfseries 0.5817 & \bfseries 1.1472 & \bfseries 2.2857  &  
                  0.1796           & 0.1767           & 0.1844            &  
                  \bfseries 0.4407 & 0.3976           & 0.3014               
        \\
        & Mixed   & 0.3249           & 0.6020           & 1.0759            &  
                  0.5929           & 1.1670           & 2.3639            &  
                  \bfseries 0.1765 & 0.1791           & 0.1972            &  
                  0.4065           & 0.3723           & 0.2904               
        \\
        \bottomrule
    \end{tabularx}
    \caption[Object and Time Specific minADE Performance]
    {   Object specific performance comparison for all metrics in
        the prediction horizon of 3s, 5s and 8s. Benchmark dataset is the validation set of Waymo Open
        Motion Dataset.}
    \label{tab:all_time}
\end{table*}

\begin{table}
    \vspace{1.5mm}
    \begin{tabularx}{\linewidth}{@{\hspace{3pt}}llcccc}
        \toprule
               Obj. Class & Model & {minADE $\downarrow$} & 
                  {minFDE $\downarrow$} & 
                  {MR $\downarrow$} & 
                  {mAP $\uparrow$}  \\ 
        \midrule
        Vehicle & Static  & 0.7543           & 1.5222           & 0.1528           & \bfseries 0.4543 \\
            &Dynamic & 0.7800           & 1.5331           & \bfseries 0.1504 & 0.4231           \\
            &Mixed   & \bfseries 0.7521 & \bfseries 1.4922 & 0.1518           & 0.4475           \\
        \midrule
        Pedestrian 
            &Static  & 0.3449           & 0.7230           & 0.0728           & 0.4142           \\
            &Dynamic & \bfseries 0.3435 & \bfseries 0.7127 & \bfseries 0.0691 & 0.4432           \\
            &Mixed   & 0.3489           & 0.7218           & \bfseries 0.0691 & \bfseries 0.4552 \\
        \midrule
        Cyclist  
            &Static  & 0.7071           & 1.4323           & \bfseries 0.1749 & \bfseries 0.3891 \\
            &Dynamic & \bfseries 0.6534 & \bfseries 1.3382 & 0.1802           & 0.3799           \\
            &Mixed   & 0.6676           & 1.3746           & 0.1843           & 0.3564           \\
        \bottomrule
    \end{tabularx}
    \caption{Object specific performance comparison of motion prediction on the validation set of Waymo Open Motion Dataset.}
    \label{tab:metric_object}
\end{table}

At a first glance at the overall metrics in \cref{tab:metric_overall} we can see that there are improvements for the dynamic and mixed model compared to the static baseline. Especially in the minADE and minFDE metrics. We decided to split the results into the respective object classes since only the vehicle object class uses dynamic intention points and the pedestrian and cyclist object classes still use the static intention points (see \cref{tab:metric_object}). Furthermore, we evaluate the results at different points in time. Namely at 3s, 5s, and 8s as can be seen in \cref{tab:all_time}. 

We observed a significant impact on the pedestrian and cyclist predictions when using the dynamic and mixed intention points for the vehicle class. There was a 7\% and 10\% improvement in mAP for pedestrians using the dynamic and mixed intention points respectively. For the cyclists, we observed a 7\% and 4\% reduction in minFDE for the dynamic and mixed models respectively.

Based on these results we draw the following conclusion:
There are significant cross effects between the different object classes. By changing the intention points for the vehicle class we also influence the predictions for the other object types. We attribute this behavior to the accuracy of the intention points. For vehicles, the static points are all over the place and the network learns to rely more on the map data for the actual predictions. The static intention points are only used as rough guidance, whereas the dynamic intention points provide much more accurate guidance and less has to be inferred from the map. This behavior also translates to pedestrians who have more freedom in their movement and are less reliant on the map. Therefore, if static intention points force a more map based prediction this could negatively impact pedestrian predictions which are far less bound by the map. This also translates to cyclists to a lesser extent. A possible solution to remove these cross effects would be a heterogeneous network architecture in which the different object classes are handled by different encoders and decoders.

Another observation is that the impact of the non-static intents is more pronounced for the vehicle class towards the end of the 8s prediction and not so much in earlier intervals. For example, in the minFDE metric, the static model outperforms our dynamic and mixed models at 3s. But at 5s the mixed model outperforms the static model and at 8s the dynamic and the mixed model outperform the static model by a 7cm and 9cm margin respectively.
This behavior is likely explained by the dominant role of a vehicle’s immediate kinematic state (orientation and momentum) over road layout in short-term predictions. While the road layout is a key factor in dynamic intention point generation that shapes long-term trajectories, its influence diminishes in brief time horizons, where motion dynamics prevail.
The more informed dynamic intention points provide additional information at longer time horizons leading to an improved performance in those scenarios.

\subsection{Qualitative assessment}
\begin{figure}
    \vspace{1.5mm}
    \begin{minipage}{0.975\linewidth}
        \centering
        \subfloat[]{\label{fig:dyn_worse_1}\fbox{\includegraphics[trim=600 600 50 550, clip,width=1\columnwidth]{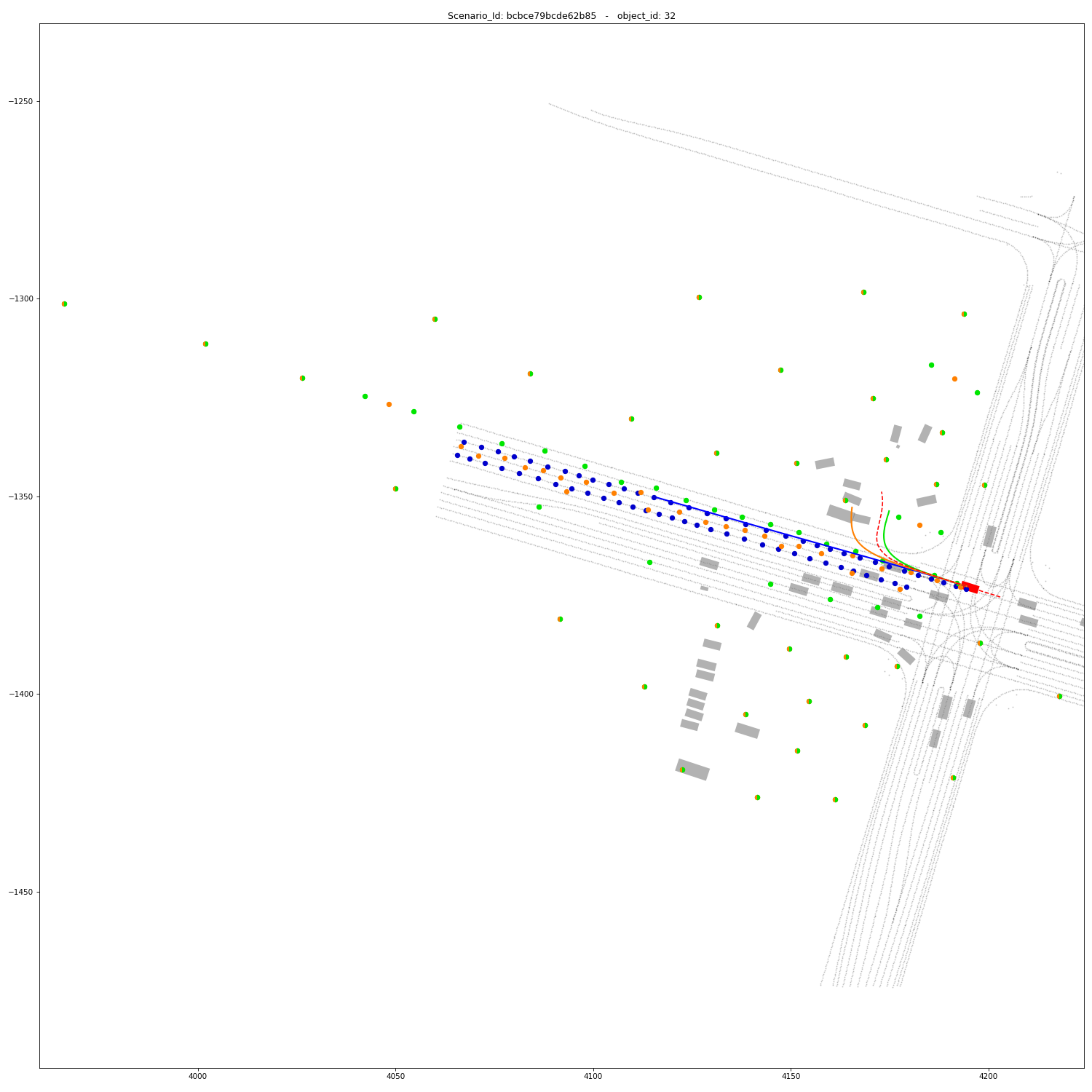}}}
    \end{minipage}%
    \vspace{1mm}
    \\
    \begin{minipage}{.45\linewidth}
        \centering
        \subfloat[]{\label{dyn_worse_2:a}\fbox{\includegraphics[trim=500 900 500 100, clip,width=\textwidth]{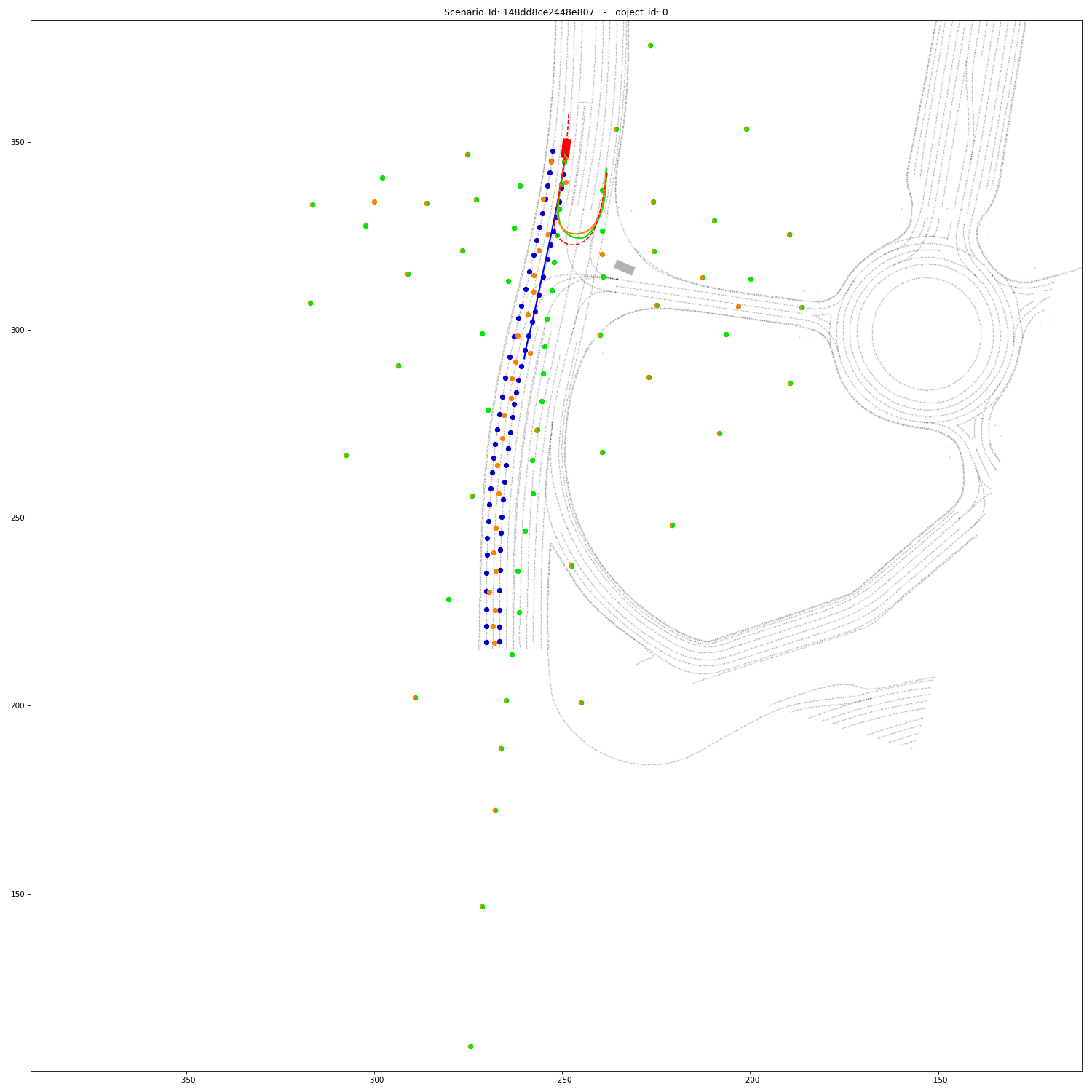}}}
    \end{minipage}%
    \hspace{6.5mm}%
    \begin{minipage}{.45\linewidth}
        \centering
        \subfloat[]{\label{dyn_worse_2:b}\fbox{\includegraphics[trim=900 525 150 525, clip,width=\textwidth]{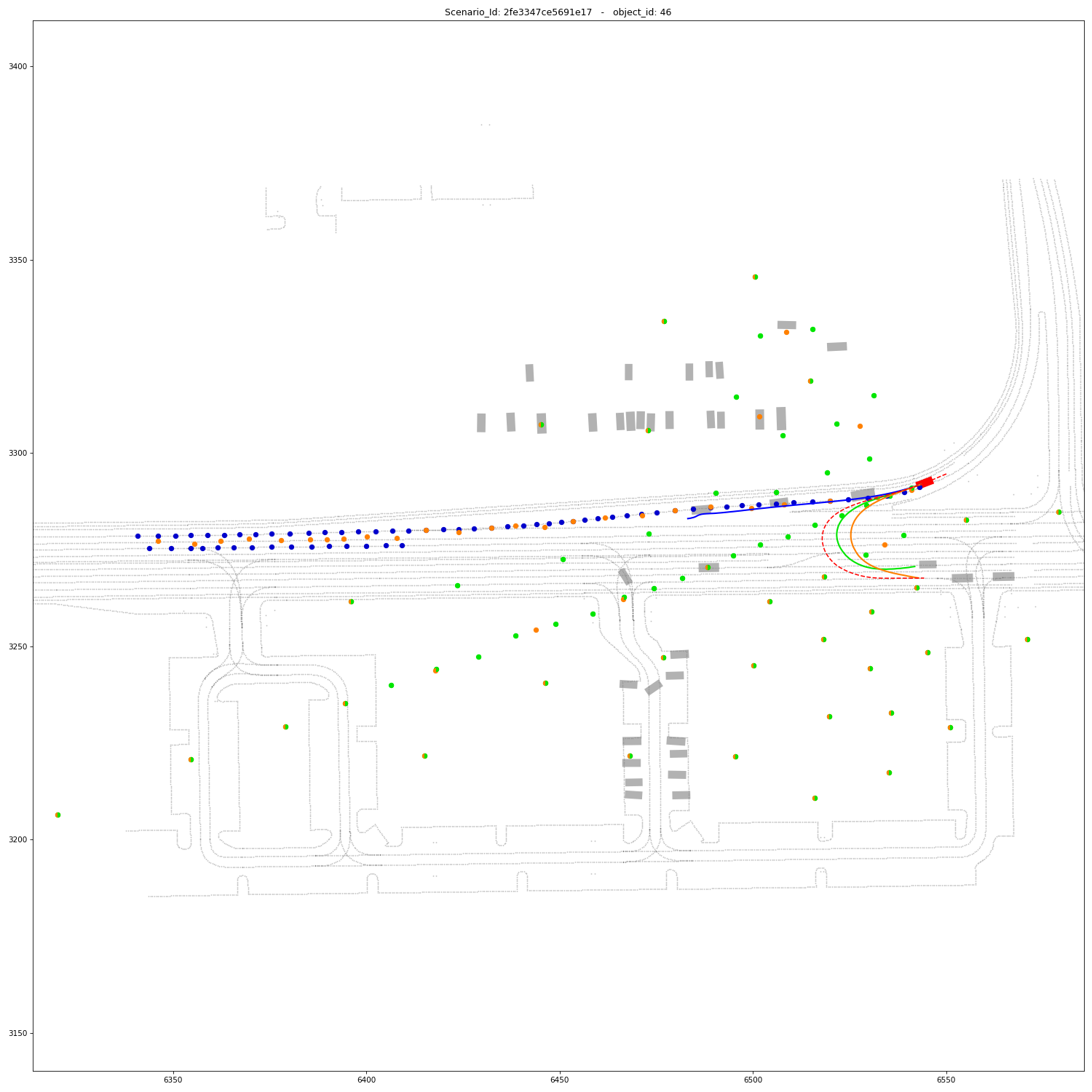}}}
    \end{minipage}
    \caption
    {Different illegal or non map conform scenarios in which the dynamic model struggles. However, the mixed and static models manage to predict these maneuvers.
        The red path denotes the ground-truth trajectory.
        Predictions from the static, dynamic, and mixed models are
        illustrated in green, blue, and orange, respectively.}
    \label{fig:dyn_worse_2}
    \vspace{-2.5mm}
\end{figure}

In our qualitative analysis, we examine various scenarios that showcase the distinct
capabilities of the different trained models.
One category of interesting scenes includes movements that are not captured by the dynamic
intention points. Illegal maneuvers, such as disregarding traffic rules or errors
in the HD map, can significantly affect the predictive performance of the dynamic model.
In \cref{fig:dyn_worse_1}, a scenario is depicted where an agent departs from the road to
enter an unmarked parking lot. In \cref{dyn_worse_2:a} an agent performs an illegal U-turn. In \cref{dyn_worse_2:b} an agent is driving on a lane that merges into another lane going the same direction. Instead of following this lane the agent aggressively cuts across multiple lanes into the oncoming traffic. In all of these instances, the maneuvers are
not represented in the road graph and are thus not included in the dynamic intention point generation.
However, the mixed model, benefiting from the incorporation of static intents, can anticipate these moves.

\subsection{Legal / map-conform agent movement}
As seen in the qualitative assessment, compared to the static and mixed model, the dynamic model struggles with illegal or non map-conform movements. We want to further evaluate the impact of such scenarios on the performance of the dynamic model.
A significant challenge in this evaluation is the
categorization of movements into 'legal' and 'illegal,' especially given the lack of
explicit legality labels in the dataset.

To address this, we hypothesize that most illegal movements
would likely deviate from the road graph we generate for the dynamic intents.
Since the dynamic intentions rely on the generated road graph, it should affect the prediction performance for illegal moves negatively.
To ensure the relevancy of our analysis, we narrowed down the validation dataset
to vehicle predictions that are suitable for generating dynamic intention points.
We excluded scenarios where generating such points was infeasible, like vehicles in
parking lots far from the road. This exclusion and the focus on vehicles reduced the
dataset from 192,172 elements to 160,005, omitting 26,496 elements involving
pedestrians or cyclists and 5,675 vehicle predictions unsuitable for dynamic intention generation. Additionally, we excluded all trajectories with missing or implausible ground truth trajectory values at the relevant timesteps. This left us with 121,930 valid trajectories.

\begin{figure}
    \vspace{1.5mm}
    \subfloat[]{\label{fig:illegal_moves:a}\includegraphics[width=0.98\columnwidth]{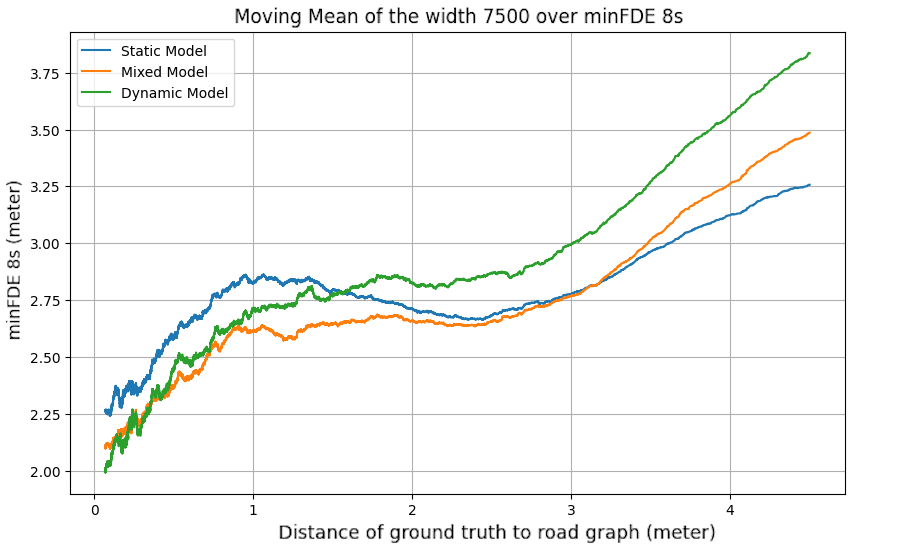}}
    \vspace{1mm}
    \subfloat[]{\label{fig:illegal_moves:b}
        \includegraphics[width=0.98\columnwidth]{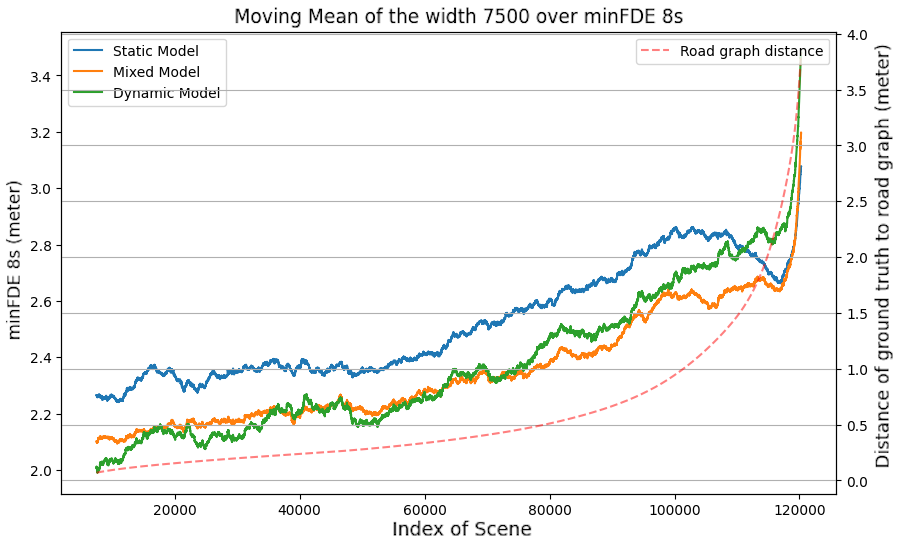}}
        \caption[Model Performance Evaluated by the minFDE 8s Metric in Relation to
        the Road Graph Distance]
    {Graph showing the correlation between the minFDE~8s score and the deviation between the ground truth trajectory endpoint and the roadgraph for all legal agent movements. The x-axis in (a) displays the deviation in meters. In (b) the x-axis shows the index of the scenarios sorted by their deviation. The actual deviation is shown by the red line.}
    \label{fig:illegal_moves}
    \vspace{-1.5mm}
\end{figure}

We focused on the minFDE over 8s as a key metric, assuming that illegal movements
deviating significantly from the anticipated direction would manifest the largest discrepancies
in this metric.
To enhance the precision of our analysis and reduce statistical noise,
we implemented a moving average with a sliding window size of 7,500 prediction scenarios.
Finally, we determine the smallest deviation between the road graph for legal agent movement and the trajectory endpoint of the ground truth trajectory for each prediction
scenario. The minFDE 8s metric was plotted against this deviation, as illustrated in \cref{fig:illegal_moves:a}.

Our findings reveal that both of our approaches, the dynamic and mixed model, outperform the static model up to a
deviation of 1m between the ground truth endpoint and the road graph, with a notable discrepancy of approximately 20cm.
Between 1m and 1.5m deviation, the dynamic model's advantage diminishes, and beyond that,
its performance progressively worsens compared to the static model.
The mixed model continues to show a better performance than the dynamic and static model up to 3m deviation. The original static model is only in these edge cases over 3m better than the mixed model. 

One interesting observation is that there is a noticeable performance improvement for the static model at the \mbox{2m\ -\ 2.5m} deviation range compared to predictions for scenarios with less deviation. This indicates that there are many easier predictions in this range. A logical explanation for this is parked cars at the side of the road which are not moving and therefore easier to predict. They fit in this deviation range. A decent autonomous driving software stack is able to detect parked vehicles without requiring trajectory predictions. When excluding parked vehicles we found that the positive impact of our dynamic and mixed models is further increased compared to the static approach.

In general, we can say that our dynamic and mixed models perform better for legal and map conform scenarios. As can be seen in \cref{fig:illegal_moves:b}, scenarios with larger deviations are very uncommon in the dataset. Only in these edge cases, the static model performs better. 
A prediction model integrated into a fully autonomous driving software stack must avoid inaccurately predicting extreme maneuvers that deviate from map data. If such false predictions occur, the autonomous vehicle may respond in ways that compromise safety and passenger comfort. To address this, the software stack could include a dedicated module to identify and handle illegal maneuvers, thereby reducing the need for the main trajectory prediction module to account for these edge cases.
In such a software stack, we expect our dynamic and mixed models to perform significantly better than the static model.
\section{Conclusion}
We evaluated the impact of dynamic and mixed intention points on the MTR model which originally only uses scene independent static intention points. Our findings show that dynamic and mixed intention points provide improved performance, especially in map conform scenarios which represent the vast majority of traffic situations. 
Another notable finding is that there are significant unintentional cross-effects between the predictions for different object classes. By switching to dynamic and mixed intention points for the vehicle class we observed an improved performance of respectively 7\% and 10\% mAP for the pedestrians which use the original static intention points. These findings imply that a future heterogeneous network architecture can improve performance by breaking these unintentional cross-effects.










{\small
\bibliographystyle{IEEEtran}
\bibliography{main}
}

\end{document}